\documentclass[letterpaper]{article}
\usepackage{authblk}
\usepackage{iccc}

\usepackage{todonotes}
\usepackage{makecell}

\usepackage{times}
\usepackage{helvet}
\usepackage{courier}

\title{Bits of Grass: Does GPT already know how to write like Whitman?
}

\author{
Piotr Sawicki\textsuperscript{1}, Marek Grze\'{s}\textsuperscript{1}, Fabricio Goes\textsuperscript{3}, Dan Brown\textsuperscript{2}, Max Peeperkorn\textsuperscript{1}, Aisha Khatun\textsuperscript{2}
\\
\textsuperscript{1 }School of Computing, University of Kent, Canterbury, UK\\
\textsuperscript{2 }Cheriton School of Computer Science, University of Waterloo, Canada\\
\textsuperscript{3 }Computing and Mathematical Sciences Department, University of Leicester, UK\\ 
P.Sawicki@kent.ac.uk, M.Grzes@kent.ac.uk, fabricio.goes@leicester.ac.uk, \\Dan.Brown@uwaterloo.ca, m.peeperkorn@kent.ac.uk, aisha.khatun@uwaterloo.ca\\
}

\begin{document} 

\maketitle

\begin{abstract}
This study examines the ability of GPT-3.5, GPT-3.5-turbo (ChatGPT) and GPT-4 models to generate poems in the style of specific authors using zero-shot and many-shot prompts (which use the maximum  context length of 8192 tokens). We assess the performance of models that are not fine-tuned for generating poetry in the style of specific authors, via automated evaluation. Our findings indicate that without fine-tuning, even when provided with the maximum number of 17 poem examples (8192 tokens) in the prompt, these models do not generate poetry in the desired style.
\end{abstract}

\section{Introduction}
The recently introduced GPT-3.5 and GPT-4 models represent significant progress over the previous versions of GPT models, achieving human-like performance on many tasks that were so far unattainable to Large Language Models (LLMs)~\cite{openai2023gpt4,bubeck2023sparks}. Among creatives tasks, GPT models can write poetry~\cite{gwerngpt3}. In this study, however, we are concerned with generating poetry in the styles of specific authors. In our previous work~\cite{sawicki2023onthepower} we have examined examined generating poetry in the style of  specific authors through fine-tuning GPT-3, with notable success. We have found that poetry generated from GPT-3.5 (text-davinci-003) through prompt engineering alone does not produce  poetry that follows the style of the requested author. In here, our aim is to investigate this finding further and also to check whether GPT-3.5-turbo (ChatGPT) or GPT-4 can achieve this task through prompting only. To facilitate comparison with the above-mentioned work, we attempt to generate poetry in the style of Walt Whitman without prior fine-tuning of the GPT models.

For that purpose, we conduct a number of experiments where we generate poetry in the style of Walt Whitman using prompts only, and we evaluate these poems against the original works of Whitman using the automated evaluation workflows presented in our previous works~\cite{sawicki2022training,sawicki2023onthepower}. 

As a main contribution of this paper, we demonstrate that generating poetry in the style of a specific author through prompting alone (whether with zero-shot or many-shot) from GPT-3.5, GPT-3.5-turbo (ChatGPT) and GPT-4 does not produce good outcome, and therefore fine-tuning is still the recommended approach. 

In the next section, we describe our experimental setup, which includes three experiments to address our research questions. In the last section, we summarize the findings of this paper and suggest the directions for future work.

\section{Method}

In this section, we describe the methodology used in this paper. First, we visually compare the difference between poems generated through the same prompt from consecutive GPT models. Then, we examine whether GPT is able to retrieve the original poems by Whitman. After that, we describe the data used for further experiments, the evaluation process, and our core results.

\subsection{Four Different Models And One Prompt}
While experimenting with poetry generation from consecutive versions of GPT, we have observed that the models produce poems of increasing level of complexity and length; however, the requested style is clearly not preserved. For example, Walt Whitman's poetry does not follow the `four lines in a stanza' structure, and does not use rhyming~\cite{bohan1995}. The majority of poems that we generated `in the style of Walt Whitman' do follow the `four lines in a stanza' structure and use rhyming. This, in fact, applies to most poetry generated from GPT models (including GPT-4). Only rarely will GPT deviate from this specific structure, and even then, the style does not match that of the requested author. This applies both to zero-shot prompting (where the prompt contains only the instruction to write a poem in the style of the specific author)  and few-shot prompting (where in the prompt, apart from the instruction, we provide as examples a few poems by the original author).  For that matter, even in a multi-step conversation with ChatGPT (GPT-3.5-turbo) and GPT-4, when the prompt highlights that the generated poems have been in 4-line stanzas with rhyme, and that the desired output should not have this structure, the model, for the most of time, still generates 4-line stanzas with rhyme. 

Table~\ref{tab:poem1} shows examples of the first stanzas of the poems generated from the same prompt by all three GPT models we are examining, followed by a fragment of a poem generated from the same prompt using GPT-3 Curie fine-tuned for 4 epochs in our previous work~\cite{sawicki2023onthepower} on the poetry of Walt Withman. There is an obvious contrast between the style of writing of the fine-tuned model and the default models.

\begin{table}[]

    \centering
    \begin{tabular}{|l|}
    \hline \\
\begin{minipage}{3.3in}

PROMPT:
\vspace{1em}

Write a poem in the style of Walt Whitman.

The poem should match the following summary:
\vspace{1em}

This is a poem about having courage and strength in the face of difficulties and hardships. The speaker is confident in their own determination and power, regardless of what life throws their way. They do not fear the darkness or horrors of life, knowing that they have the strength to handle what comes their way.
\vspace{1em}

GPT-3.5:
\vspace{1em}

Oh how I ride and stride,

In the darkest night and night's surprise.

What trouble can come my way, 

Will not break me, no matter how I dread.
\vspace{1em}

GPT-3.5-turbo: 
\vspace{1em}

Oh, the trials and tribulations of life,

How they test our will and might,

But fear not, my dear friend,

For courage and strength are within sight.

\vspace{1em}

GPT-4:
\vspace{1em}

I sing the song of the sailor, the wanderer of the seas,

The one who beholds the world with eyes wide and free,

In the port of Charleston, where the ships lay anchored,

I saw a sight that stirred my soul, a vision that lingered.

\vspace{1em}

Fine-tuned (4 epochs) GPT-3 Curie:
\vspace{1em}

OUT of the night that covers me like a hearse,

Ambient and sorrowful as the murmurous verse of those I love best,

(Not the unreplying march of Death, whose cold speech is wholly chang’d,

But the reply, the matin song of all that is, and ever shall be, 

in spire, in choir,

Out of the sung response, arriere glissade, in the ranks of the

living,)

A vibration, blue and tremulous as a cloud of wakeful birds,

A departure and a procession stealing away into the night,

(...)

\vspace{1em}

\vspace{1mm}
\end{minipage}
\\ \hline
    \end{tabular}
    \caption{Example of the first stanzas of the poems generated by GPT-3.5, GPT-3.5-turbo and GPT-4 from the prompt shown, followed by the fragment of a poem generated from the same prompt from fine-tuned GPT-3 Curie. The summary provided in the prompt is of the poem `Out Of The Night That Covers Me' by William Ernest Henley. For the fine-tuned model, only the summary without the verbal instruction was entered as a prompt.}
    \label{tab:poem1}
\end{table}

\begin{table}[t]
\begin{center}
{\small
\begin{tabular}{ |l|c|c|l| }
\hline
\multicolumn{4}{|c|}{Retrieving complete text of Whitman's poems} \\
\hline
Poem title & GPT-3.5 & GPT-3.5-turbo & GPT-4\\
\hline
\makecell[l]{Spirit Whose\\Work Is Done} & 24.60\% & 96.05\% & 20.68\% \\
\hline
\makecell[l]{Aboard At\\A Ship's Helm} & 26.43\% & 91.96\% & 94.79\% \\
\hline
\makecell[l]{Who Learns My\\Lesson Complete?} & 21.21\% & 16.09\% & 49.59\% \\
\hline
\makecell[l]{The World Below\\the Brine} & 28.06\% & 98.53\% & 98.53\% \\
\hline
\makecell[l]{As At Thy Portals\\Also Death} & 27.16\% & 99.47\% & 99.47\% \\
\hline
Eidólons & 15.19\% & 13.82\% & 94.42\% \\
\hline
\makecell[l]{I was Looking\\ a Long While} & 27.60\% & 98.02\% & 98.14\% \\
\hline
\makecell[l]{Italian Music in\\ Dakota} & 24.34\% &\textbf{0.0\%} & 82.28\% \\
\hline
Miracles & 22.81\% & 45.31\% & 67.18\% \\
\hline
\makecell[l]{By Broad Potomac's\\Shore} & 25.05\% & 24.34\% & 23.66\% \\
\hline

\hline
\textbf{Avg. Result} & \textbf{24.25\%} & \textbf{58.36\%} & \textbf{72.87\%} \\
\hline
\end{tabular}
}
\end{center}
\vspace{-10px}
\caption{Results of retrieving the complete text of the poems by our chosen author. The average Levenshtein distance, calculated over five trials, is utilized to quantify the similarity between the retrieved text and the original poems.}\label{tbl:retrieve}
\vspace{-10px}
\end{table}

\subsection{Does GPT Know Whitman's Poems?}

Before proceeding to poetry generation and evaluation, we first wanted to examine whether GPT is `aware' of the style of writing we are requesting. For that, we have run a simple experiment to check the GPT's ability to provide the complete text  of a requested poem.
We randomly selected 10 poems by Walt Whitman, and asked each of the tested GPT models to retrieve the text of the poems using the following prompt:
\begin{footnotesize}
\begin{verbatim}
Give me the text of a poem
{TITLE OF THE POEM} by Walt Whitman.
\end{verbatim}
\end{footnotesize}

Unlike in the previous versions of GPT, in GPT-3.5-turbo and GPT-4, setting the temperature parameter to 0 does not guarantee repeatability. For this reason, the process was repeated 5 times for every poem and the results were averaged. The averaged results are shown in Table~\ref{tbl:retrieve}. The similarity score reported is Levenshtein distance~\cite{levenshtein1966binary} between the original poem and the poem retrieved by the model. The Levenshtein distance is an efficient and versatile method for measuring string similarity, as it determines the minimal number of single-character edits needed to convert one string into another.

The results above 90\% indicate correctly retrieved poems, with some minor differences in layout. This is acceptable, since these kind of differences are found even between different websites presenting the same poem. The lower results on GPT-3.5-turbo and GPT-4 almost always indicate that the models started to retrieve the poem correctly, but than deviated from the original text. However, the GPT-3.5 model has never correctly retrieved even a fragment of a requested poem, and we suspect that in the case of this model the results are always around 20\% because of similar vocabulary used. It is interesting to note that in the case of ``Italian Music in Dakota", GPT-3.5-turbo in all five attempts have responded: \emph{`I'm sorry, but Walt Whitman did not write a poem titled ``Italian Music in Dakota. It is possible that you are thinking of a different poet or a different poem title.'}. Therefore, we have entered 0.0\% for this poem.

We can speculate that GPT's ability to retrieve the text of the poems is influenced by the number of times the poem appeared in the training dataset. Regardless, GPT-3.5-turbo and GPT-4 are, in many cases, able to retrieve the requested poems, and therefore, we can assume that GPT-3.5-turbo and GPT-4 ``know'' the style of this poet.

\subsection{Experimental Setup}
The principal focus of this paper is on evaluating the poetry generated through zero-shot prompts.
In Reynolds and McDonell~\shortcite{reynolds2021prompt} it is argued that few-shot prompting is in many cases unnecessary. For example, in translation: it is not reasonable to assume that the language models can learn to translate from language A to language B just from the few examples provided in the few-shot prompt. Those works argue that the LLM already possesses the skill of (for example) translating between the two given languages, and the only purpose of the prompt is to `invoke' that particular skill. We speculate that this argument could extend to poetry generation using LLMs.

We were, however, intrigued by the possibility of using 8192 token-long prompts in the current version of GPT-4, which was launched 7 weeks before the submission deadline for this paper. Therefore, we also include a preliminary evaluation of poems generated from maximum-length many-shot prompts.

\begin{table}[]
    \centering
    \renewcommand{\arraystretch}{1.2}
    \begin{tabular}{|l|l|}
    \hline Model & Version \\ \hline
GPT-3.5 & text-davinci-003  \\ \hline
ChatGPT & gpt-3.5-turbo (v. 2023.04.08) \\ \hline
GPT-4 & gpt-4 (v. 2023.04.08) \\ \hline

    \end{tabular}
    \vspace{-5px}
    \caption{GPT versions used for poetry generation.}
    \label{TAB:gpt_versions}
    \vspace{-15px}
\end{table}

\subsection{Data Preparation}
The original author we have chosen for this work is Walt Whitman (American, 1819–1892). We use the dataset of his works created in our previous work~\cite{sawicki2023onthepower} that is available on our GitHub repository\footnote{https://github.com/PeterS111/Fine-tuning-GPT-3-for-Poetry-Generation-and-Evaluation}, which contains 300 poems for seven different authors (including Whitman). Since we are examining all three of the top GPT models: GPT-3.5, GPT-3.5-turbo and GPT-4 (Table~\ref{TAB:gpt_versions}) with zero-shot prompting, and additionally we are examining GPT-4 with many-shot prompting, we have prepared four datasets to be used in this experiment. To match the 300 samples of the original author's works, we generate 300 samples from each of the GPT models examined, using the following prompt for zero-shot prompting:
\begin{footnotesize}
\begin{verbatim}
Write a poem in the style of {AUTHOR}. 
The poem should match the following summary:
{SUMMARY OF THE POEM}
\end{verbatim}
\end{footnotesize}
We experimented with different ways of structuring the zero-shot prompts, but have found no meaningful differences in output quality between them.

In the case of many-shot prompting of GPT-4, we generated 300 samples with the maximum possible prompt length (8192 tokens), where, apart from the instruction to generate the poem, we provided as examples 17 poems by Whitman accompanied by their summaries. The poems included in the 17-shot (i.e. 17-poem) prompt are the following: `1861', `A Woman Waits For Me', `Spain 1873-'74', `Sparkles From The Wheel', `Spirit Whose Work Is Done', `States!', `Tears', `That Music Always Round Me', `The Artilleryman's Vision', `The Base Of All Metaphysics', `The City Dead-House', `The Indications', `Aboard At A Ship's Helm', `The Ox tamer', `The World Below The Brine', `These, I, Singing In Spring', and `Think Of The Soul'. The structure of the `17-poem' prompt is as follows:
\begin{footnotesize}
\begin{verbatim}
These are the examples of prompts and 
completions. Prompt contains the summary 
of the poem, completions contains the poem 
based on this summary. Write the last 
completion from the prompt preceeding it, 
following the examples given.
PROMPT:
{SUMMARY OF POEM 1}
COMPLETION:
{BODY OF POEM 1}
......
PROMPT:
{SUMMARY OF POEM 17}
COMPLETION:
{BODY OF POEM 17}
PROMPT: 
{SUMMARY OF THE POEM TO BE GENERATED,
FROM HENLEY AND ROSETTI DATASET}
COMPLETION:
\end{verbatim}
\end{footnotesize}

As before, we experimented with various ways of structuring this prompt, but found no significant differences in the output quality. One of the approaches we tried was to provide the 17-poem prompt shown above, but without the verbal instruction preceding it, thus attempting to simulate the fine-tuning process, but that did not improve the output quality.

The summaries we use for our poem generation (both zero-shot and many-shot) are taken from our dataset published in~\cite{sawicki2023onthepower}, and these are the same summaries that were used by us for poetry generation from their fine-tuned models. These summaries were generated for poems by William Ernest Henley (1849–1903) and Christina Rossetti (1830–1894). There are 150 summaries for each author, giving 300 summaries in total. Overall, we obtain four datasets, each containing 300 poems generated from a specific GPT model as label 0, and 300 poems by the original author as label 1. Each dataset is split into training/validation subsets, with 200/100 samples per label, respectively. This two-label setup is necessary for evaluation with binary classifiers described in the Evaluation section.

When examining the dataset generated from the 17-poem prompts, we have observed that only about 25\% of generated poems have deviated from the structured/rhymed style and on the surface have resembled Whitman's poetry. We can speculate that the model produces `higher quality' outputs when prompted with a summary which is related to the subject that Whitman was writing about, and fails when we request a poem on the subject that is not present in Whitman's works, but that would require detailed analysis by the expert in English literature.

We have to stress that few-shot and many-shot prompting of GPT-4 requires a dedicated study, and in here it was treated only as a preliminary experiment.

\subsection{Evaluation}

Having prepared the datasets, we are fine-tuning GPT-3 for binary classification, following the automated evaluation methodology presented in our previous work~\cite{sawicki2023onthepower}, where evaluation is done in the following way: binary classifiers are trained on two labels, label 0 being the  GPT output, and label 1 the works of the original author. If the classifier cannot distinguish between those two classes, it means that the generated poems are of `good' quality. On the contrary, if the classifier can distinguish between the two classes, it means that generated poems do not match the style/quality of the original author. Achieving a 50\% score would mean that both labels are indistinguishable to our classifiers, which is the desired outcome.

Following the findings in~\cite{sawicki2023onthepower}, we have chosen the GPT-3 Babbage as a basis for fine-tuning the classifiers. The results of classification for both authors are shown in Table~\ref{tbl:exp_1_results}. It additionally includes the results from the best performing fine-tuned GPT-3 model for Whitman's poetry (FT-GPT-3 Curie 4 epochs) from~\cite{sawicki2023onthepower}. We can compare our fine-tuned models' results with the current results because of the matching setup, i.e., we used the same dataset of Whitman's works, our evaluation setup contained the same amount of samples per label, the training/evaluation split was the same (200/100), and the poems were generated from the same set of summaries.

The results show that the classifiers were able to distinguish the GPT-generated poems from the original authors' works with almost 100\% accuracy. This shows that the poems generated through prompting only do not match the style/quality of writing of the original authors, while the poems generated from the fine-tuned GPT-3 models~\cite{sawicki2023onthepower} are approaching the style/quality of the original authors' works. 

These results should be interpreted with caution in the light of the fact that the binary classifiers used are entirely black-box systems, i.e. we do not know how the classification was performed. However, knowing that fine-tuned GPT-3 models are reliable as binary classifiers, as shown in~\cite{sawicki2023onthepower}, we can, to some extent, rely on these results. Further investigation, especially including human evaluations, is necessary to thoroughly determine the quality of the GPT-generated poetry.

\begin{table}[t]
\begin{center}
    \renewcommand{\arraystretch}{1.2}
{\small
\begin{tabular}{ |l|c|c|l| }
\hline
\multicolumn{4}{|c|}{GPT-x vs Walt Whitman original} \\
\hline
Model & Correct & Incorrect & Accuracy\\
\hline
GPT-3.5 & 200 & 0 & 100\% \\
\hline
GPT-3.5-turbo & 200 & 0 & 100\% \\
\hline
GPT-4 & 200 & 0 & 100\% \\
\hline
GPT-4 17-poem prompt & 199 & 1 & 99.5\% \\
\hline
\textbf{FT-GPT-3 Curie 4e} & \textbf{123} & \textbf{77} & \textbf{61.5\%} \\
\hline

\end{tabular}
}
\end{center}
\vspace{-10px}
\caption{Results of our experiments where GPT-generated poetry is compared against the Walt Whitman's original works. Entries in the first column indicate which GPT model's output was evaluated against the Whitman's works.}\label{tbl:exp_1_results}
\vspace{-10px}
\end{table}

\section{Conclusion}

In this study, we have examined the poetry generation ability of GPT-3.5, GPT-3.5-turbo and GPT-4 when used with prompting only. We have found that the generated poems do not match the style/quality of the works of the original author, whereas the fine-tuned model can consistently reproduce the complex style of an author like Whitman.  It remains to be seen whether later versions of GPT will render the fine-tuning process obsolete (for the purpose of generating poetry in the style of a specific author), but as of now, using prompting of default GPT models does not produce good results, and fine-tuning is a recommended approach. 

\section{Acknowledgments} 
The work of DB is supported by a Discovery Grant from the Natural Sciences and Engineering Council of Canada. MP is supported by the University of Kent GTA Studentship Award, Prins Bernhard Cultuurfonds, Hendrik Mullerfonds, and Vreedefonds.

\section{Author contributions}
Experimental design: PS with MG, FG, DB, MP; experimental implementation: PS; writing: PS with MG, DB, FG, editing: MG, DB, FG, MP, AK.

\bibliographystyle{iccc}
\bibliography{iccc}

\end{document}